\documentclass[11pt]{article}

\usepackage[margin=1in]{geometry}
\usepackage[T1]{fontenc}
\usepackage[utf8]{inputenc}
\usepackage{lmodern}
\usepackage{microtype}

\usepackage{amsmath,amssymb}
\usepackage{booktabs}
\usepackage{hyperref}
\usepackage{graphicx}
\usepackage{enumitem}
\usepackage{tikz}
\usetikzlibrary{arrows.meta,positioning,shapes.geometric}

\title{Bridging Philosophy and Machine Learning:\\
A Structuralist Framework for Classifying Neural Network Representations}

\author{%
  Yildiz Culcu\\[4pt]
  \small Department of Computer Science, Christian-Albrechts-Universität zu Kiel\\
  \small Max Planck Institute for Cognitive and Brain Sciences\\
}

\date{} 

\begin{document}

\maketitle

\begin{abstract}
Machine learning models increasingly function as representational systems, yet the philosophical assumptions underlying their internal structures remain largely unexamined. This paper develops a structuralist decision framework for classifying the implicit ontological commitments made in machine learning research on neural network representations. Using a modified PRISMA protocol, a systematic review of the last two decades of literature on representation learning and interpretability is conducted. Five influential papers are analysed through three hierarchical criteria derived from structuralist philosophy of science: entity elimination, source of structure, and mode of existence. The results reveal a pronounced tendency toward structural idealism, where learned representations are treated as model-dependent constructions shaped by architecture, data priors, and training dynamics. Eliminative and non-eliminative structuralist stances appear selectively, while structural realism is notably absent. The proposed framework clarifies conceptual tensions in debates on interpretability, emergence, and epistemic trust in machine learning, and offers a rigorous foundation for future interdisciplinary work between philosophy of science and machine learning.
\end{abstract}


\section{Introduction}

\subsection{The parallel evolution of representation research}

Research on ``representation'' has developed independently across multiple disciplines. In machine learning, representation learning investigates how neural networks encode information in their internal states~[1]. In philosophy of science, scholars examine how theories and models represent physical phenomena~[21]. In neuroscience, researchers study how neural activity encodes sensory information and gives rise to mental representations, directly confronting the mind--body problem~[20].

Neural networks develop internal representations through training. In vision models, early layers detect edges and textures while deeper layers recognise objects and scenes~[23]. In large language models (LLMs), neurons activate for specific concepts---some respond to geographic entities, others to temporal expressions, and still others to syntactic structures~[6]. Recent work shows that LLMs develop ``world models'', with internal states that track entities, relations, and even truth values across contexts~[12].

These representations emerge without explicit programming. When GPT-style models learn to complete text, they develop representations of grammar, factual knowledge, and reasoning patterns. When diffusion models learn to generate images, they encode artistic styles, object relationships, and scene composition. A central question arises: Are these representations discoveries of genuine patterns or useful constructions?

Philosophy of science asks fundamental questions about scientific representation. Structural realists argue that successful theories capture real structures in nature, even if concepts of objects change~[22]. Anti-realists contend that representations are instruments for prediction rather than descriptions of reality~[21]. Mathematical structuralists debate whether mathematical objects exist independently or only as positions in structures~[19].

These debates have practical implications. If neural networks discover real patterns, one should expect convergence across architectures. If they construct representations, it becomes crucial to understand the biases that these constructions introduce.

\subsection{Research gap and main question}

Despite clear parallels between philosophical questions and machine learning (ML) practice, the fields remain largely disconnected. ML researchers develop interpretability methods without examining their philosophical assumptions. Philosophers debate AI consciousness and representation without analysing actual neural network mechanisms. No systematic framework currently bridges this gap.

The main research question of this paper is:

\medskip
\noindent\textbf{Do machine learning researchers implicitly adopt structuralist philosophical positions when they discuss representations and structures in neural networks, and can these positions be systematically classified?}

\medskip

Table~\ref{tab:parallel} summarises the parallel guiding questions.

\begin{table}[t]
\centering
\caption{Parallel questions in philosophy and machine learning.}
\label{tab:parallel}
\begin{tabular}{p{3.7cm}p{3.7cm}}
\toprule
\textbf{Structuralist philosophy question} & \textbf{Machine learning parallel} \\
\midrule
Are structures discovered or constructed? &
Do networks uncover true generative factors or build useful features? \\
How are objects individuated within structures? &
How should learned features be interpreted---intrinsically or relationally? \\
Are higher-level patterns reducible? &
Are emergent capabilities reducible to network components? \\
\bottomrule
\end{tabular}
\end{table}

\section{Conceptual background}

\subsection{Methodological framework}

The study employs deductive content analysis following Mayring's framework, in which predefined theoretical categories are brought to texts as coding schemes~[15]. This strictly deductive approach uses categories established \emph{a priori} from philosophical literature---the four structuralist positions---rather than deriving categories inductively from ML papers.

A modified PRISMA (Preferred Reporting Items for Systematic Reviews and Meta-Analyses) methodology~[17] is used for transparent and reproducible literature selection. The PRISMA 2020 statement provides rigorous guidance for search strategies, inclusion and exclusion criteria, and selection processes, creating an auditable trail from database queries to final paper selection.

Content analysis follows Krippendorff's methodology for systematically analysing textual data~[10]. Unlike purely quantitative text mining or purely interpretive discourse analysis, content analysis enables systematic categorisation while preserving conceptual nuance. The framework accommodates latent content analysis---inferring underlying meanings not explicitly stated---which is essential since philosophical commitments in ML papers often remain implicit.

\subsection{Machine learning background}

Representations in machine learning refer to transformed versions of input data that capture relevant information for downstream tasks. Bengio, Courville and Vincent~[1] define representation learning as discovering transformations that make it easier to extract useful information when building classifiers or predictors. In neural networks, representations emerge at multiple granularities: individual neuron activations encode specific patterns, layer-wise activations capture hierarchical features, and the network's entire activation space forms a learned representation space.

The geometry of representation spaces has become central to understanding neural networks. Bronstein et al.\ argue that deep learning's success stems from exploiting geometric priors---symmetries, invariances and structural assumptions about data~[2]. When a neural network maps images to vectors, it creates a representation space where geometric relationships (distances, angles, clusters) correspond to semantic relationships.

Interpretability in machine learning encompasses methods for understanding model decisions and internal mechanisms. Lipton distinguishes between transparency (understanding the model itself) and post-hoc interpretability (explaining decisions after training)~[14]. The emerging field of mechanistic interpretability seeks to reverse-engineer neural networks' internal algorithms, as demonstrated by Olah et al., who argue that meaningful features might be encoded in linear combinations of neurons rather than individual units~[16].

\subsection{Philosophical foundations}

Structuralism in philosophy of science emerged from attempts to salvage scientific realism after revolutionary theory changes. Russell introduced structural realism, arguing that science reveals only structural properties of reality~[18]. Worrall's influential paper addressed the pessimistic meta-induction by distinguishing mathematical structures (preserved across theory change) from theoretical entities (frequently abandoned)~[22].

Contemporary structuralism diverged into multiple variants.

\medskip
\noindent\textbf{Definition 1 (Structural realism).} Structures exist independently in nature; science discovers them; they are instantiated in physical systems.

\medskip
\noindent\textbf{Definition 2 (Structural idealism).} Structures are human constructs imposed on experience for organisational and predictive purposes.

\medskip
\noindent\textbf{Definition 3 (Eliminative structuralism).} Only structures exist; entities are mere nodes with no independent existence.

\medskip
\noindent\textbf{Definition 4 (Non-eliminative structuralism).} Structures exist as abstract mathematical objects; entities exist but are defined by structural position.

\medskip

These positions can be mapped onto debates about neural network representations. When researchers speak of discovering generative factors, they echo structural realism. When they describe representations as artifacts of architecture and loss functions, they align with structural idealism. When interpretability work treats networks as graphs of circuits or attention patterns while downplaying individual units, it approaches eliminative structuralism. When latent variables are retained but their meaning depends on modelling choices, non-eliminative structuralism becomes salient.

\section{Methodology}

\subsection{Search strategy and selection process}

A modified PRISMA framework is employed for systematic paper selection. The search strategy utilises four databases (Google Scholar, arXiv, IEEE Xplore, ACM Digital Library) with Boolean combinations of three conceptual clusters:

\begin{enumerate}[label=(\arabic*),leftmargin=*,topsep=1pt,itemsep=1pt]
\item ML techniques: ``neural network*'', ``deep learning'', ``representation learning'';
\item Structural concepts: ``representation*'', ``structure*'', ``emergence'';
\item Philosophical dimensions: ``interpretability'', ``understanding'', ``ontology''.
\end{enumerate}

Papers were included if they: (1) were published between 2005 and 2025, (2) addressed learned representations in neural networks, (3) engaged with questions about what neural networks represent, (4) made implicit or explicit philosophical claims about structures, (5) contained substantial theoretical discussion and (6) had significant citations (approximately 20 per year since publication, adjusted by age).

The selection process followed three screening stages. Initial filtering of titles and abstracts yielded 27 papers. Abstract and citation review narrowed this to 10 papers. Full-text review and practical constraints---including the limited project timeframe and the requirement that papers be technically comprehensible without narrow subfield expertise---resulted in 5 papers for detailed analysis. This deliberate choice prioritised depth of analysis over breadth, focusing on developing and demonstrating the classification framework rather than providing comprehensive empirical coverage.

\subsection{The classification framework}

The framework employs three hierarchical criteria that translate philosophical distinctions into operational categories applicable to ML papers.

\subsubsection{Criterion 1: Entity elimination}

This criterion examines whether neural network components (neurons, features, weights, layers) are treated as possessing independent existence or conceived merely as relational nodes. The eliminative stance manifests through purely functional descriptions, graph-theoretic language and system-level evaluation. The non-eliminative approach discusses neurons or features as detectors with specific representational content, often accompanied by unit-level analysis and semantic labelling.

\subsubsection{Criterion 2: Source of structure}

This criterion investigates whether patterns and organisational principles in neural network representations are understood as discovered properties of data or world, or as structures imposed through architectural choices and training procedures. Discovery indicators include language such as ``uncover'', ``reveal'', ``discover'', and evaluation against known ground truth. Imposition indicators include verbs like ``construct'', ``design'', ``build'', and emphasis on inductive biases, architectural priors and optimisation schemes.

\subsubsection{Criterion 3: Mode of existence}

Applied only when structures are treated as discovered, this criterion examines whether structures are understood as empirical patterns tied to specific data distributions or as abstract mathematical entities. Empirical indicators include dataset-specific claims, sensitivity to distribution shift and focus on robustness and generalisation boundaries. Mathematical indicators include appeals to universal approximation theorems, domain-agnostic principles and analyses carried out in abstract function spaces or infinite-width limits.

\subsection{Decision tree structure and logic}

\begin{figure}[t]
\centering
\begin{tikzpicture}[
  node distance=1.5cm and 1.8cm,
  every node/.style={align=center, font=\small},
  decision/.style={rectangle, draw, rounded corners, fill=blue!10,
                   minimum width=2.6cm, minimum height=0.8cm},
  outcome/.style={rectangle, draw, fill=green!8,
                  minimum width=2.6cm, minimum height=0.8cm},
  arrow/.style={->, >=Stealth, thick}
]

\node[decision] (C1) {Criterion 1:\\ Entity elimination};

\node[outcome, below left=of C1, xshift=1.0cm] (ES) {Eliminative\\ structuralism};
\node[decision, below right=of C1, xshift=-1.0cm] (C2) {Criterion 2:\\ Source of structure};

\node[outcome, below left=of C2, xshift=1.0cm] (SI) {Structural\\ idealism};
\node[decision, below right=of C2, xshift=-1.0cm] (C3) {Criterion 3:\\ Mode of existence};

\node[outcome, below left=of C3, xshift=1.0cm] (SR) {Structural\\ realism};
\node[outcome, below right=of C3, xshift=-1.0cm] (NES) {Non-eliminative\\ structuralism};

\draw[arrow] (C1) -- node[left,pos=0.35]{Eliminated} (ES);
\draw[arrow] (C1) -- node[right,pos=0.35]{Retained} (C2);

\draw[arrow] (C2) -- node[left,pos=0.35]{Imposed} (SI);
\draw[arrow] (C2) -- node[right,pos=0.35]{Discovered} (C3);

\draw[arrow] (C3) -- node[left,pos=0.35]{Empirical} (SR);
\draw[arrow] (C3) -- node[right,pos=0.35]{Mathematical} (NES);

\end{tikzpicture}
\caption{Decision tree for classifying structuralist positions in neural network representation papers.}
\label{fig:decisiontree}
\end{figure}
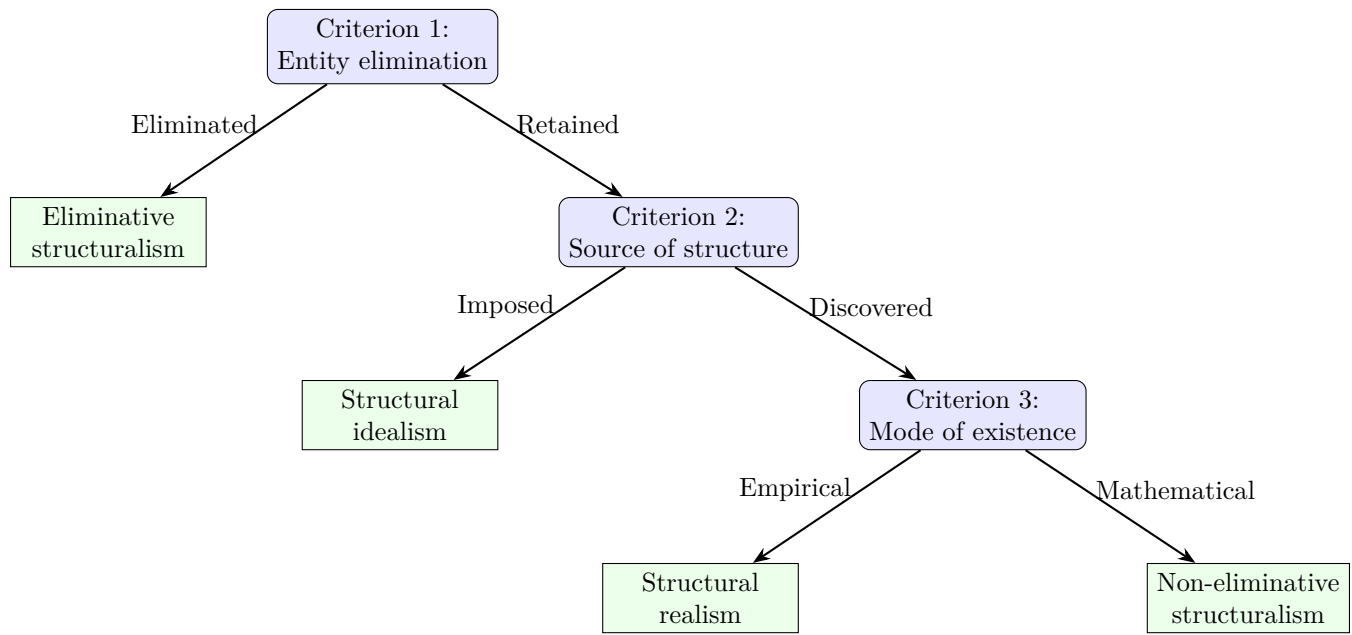

The three criteria can be represented as a decision tree, shown in Figure~\ref{fig:decisiontree}.

\section{Results}

\subsection{Paper selection and classification}

The initial search yielded 27 papers meeting the basic inclusion criteria. After the three screening phases, 5 papers were selected for detailed philosophical analysis: Lipton (2016) on interpretability, Carvalho et al.\ (2019) on interpretability methods, Hooker and Hooker (2017) on ML and realism, Hohenecker and Lukasiewicz (2020) on ontology reasoning and Heuillet et al.\ (2020) on explainable reinforcement learning.

Application of the classification framework yielded the distribution shown in Table~\ref{tab:results}.

\begin{table}[t]
\centering
\caption{Classification results for selected ML papers.}
\label{tab:results}
\begin{tabular}{lcc}
\toprule
Position & Papers & Percentage \\
\midrule
Structural idealism & 3 & 60\% \\
Non-eliminative structuralism & 1 & 20\% \\
Eliminative structuralism & 1 & 20\% \\
Structural realism & 0 & 0\% \\
\bottomrule
\end{tabular}
\end{table}

\subsection{Case analysis: Lipton's ``Mythos of Model Interpretability''}

Lipton's paper~[13] exemplifies structural idealism through consistent patterns. The paper demonstrates a non-eliminative commitment through statements such as: ``In computer vision, we frequently pass an image through multiple layers of a convolutional neural network (CNN) to generate a representation for use in classification or detection. In natural language processing \dots\ learned representations may be post-processed to provide visualisations \dots\ or to cluster together words with similar semantics'' (p.~6). This framing treats representations as semantically rich entities, not as mere graph nodes.

The imposition stance is evident in claims such as: ``The demand for interpretability arises when there is a mismatch between the formal objectives of supervised learning and the real-world goals held by the users of these models \dots\ we want interpretations to serve those objectives that we deem important but struggle to model formally'' (p.~2). Here, interpretability and, by extension, representational structure are treated as pragmatic constructions serving human purposes rather than windows onto an independent reality.

\section{Discussion}

\subsection{The predominance of structural idealism}

The empirical results reveal a consistent epistemological orientation toward structural idealism, characterised by: (1) ontological commitment to components as substantive entities, (2) a constructivist view of structures as pragmatic artefacts, (3) a pragmatic epistemology emphasising instrumental value over truth and (4) an engineering-oriented focus on practical effectiveness.

This predominance aligns with theoretical expectations. Machine learning operates at exceptionally high abstraction: parameterised transformations applied to vectorised representations that are fundamentally divorced from direct correspondence with physical entities. This abstraction gradient arguably predisposes ML researchers toward instrumentalist and constructivist epistemologies.

\subsection{An abstraction--realism inverse relationship}

The results suggest an inverse relationship between disciplinary abstraction and realist epistemological commitments. Fields characterised by high abstraction (e.g., machine learning, pure mathematics) naturally gravitate toward idealist positions. Conversely, disciplines with direct empirical access to physical systems (e.g., neuroscience, astronomy) exhibit stronger realist tendencies.

The virtual absence of structural realism in the analysed corpus is particularly illuminating. For many ML practitioners, it is unclear what it would even mean for a convolutional filter or attention weight matrix to be ``true'' in a metaphysical sense. The framework's inability to identify structural realism may thus reflect not a limitation but an accurate characterisation of the field's epistemic priorities.

\section{Limitations and future directions}

This study faces several methodological limitations. The selection process incorporates an implicit comprehension filter: papers were partially selected based on their technical accessibility to a single researcher. Time constraints limited analysis to five papers, precluding statistical inference. The single-coder design meant all classifications were performed by one analyst without formal inter-rater reliability measures.

Future research should prioritise: (1) establishing inter-rater reliability through multi-coder designs, (2) implementing fuzzy coding schemes to capture hybrid positions in single papers, (3) scaling through automation using fine-tuned language models for text classification and (4) expanding the sample to enable statistical analysis of correlations between technical methods and philosophical commitments.

\section{Conclusion}

This paper demonstrates that machine learning researchers implicitly adopt structuralist philosophical positions, with a pronounced predominance of structural idealism. The classification framework makes these hidden commitments visible and analysable, revealing that ML has overwhelmingly adopted pragmatic--instrumentalist orientations, treating neural networks as tools that construct useful representations rather than discover truths.

This is not a failure but a choice, one that aligns with ML's engineering orientation. However, as these systems increasingly shape scientific practice and social life, understanding their philosophical foundations becomes crucial. When questions are raised about whether AI systems can be aligned with human values or whether they genuinely ``understand'' anything, these are not purely technical questions with purely technical answers. They are philosophical questions that depend on what is believed about what neural networks can represent and know.

The path forward requires both methodological rigour and technological innovation. Only through scaled, systematic investigation can the philosophical landscape of machine learning be mapped, clarifying not just what the field accomplishes technically but what it assumes philosophically.

\clearpage

\section*{References}

\begin{enumerate}[leftmargin=*,itemsep=1pt,topsep=2pt]

\item Y.~Bengio, A.~Courville and P.~Vincent.
Representation learning: A review and new perspectives.
\emph{IEEE Transactions on Pattern Analysis and Machine Intelligence}, 35(8):1798--1828, 2013.

\item M.~M.~Bronstein, J.~Bruna, T.~Cohen and P.~Veličković.
Geometric deep learning: Grids, groups, graphs, geodesics, and gauges.
\emph{IEEE Signal Processing Magazine}, 38(4):42--53, 2021.

\item D.~V.~Carvalho, E.~M.~Pereira and J.~S.~Cardoso.
Machine learning interpretability: A survey on methods and metrics.
\emph{Electronics}, 8(8):832, 2019.

\item A.~Chakravartty.
\emph{A Metaphysics for Scientific Realism: Knowing the Unobservable}.
Cambridge University Press, 2007.

\item S.~French.
\emph{The Structure of the World: Metaphysics and Representation}.
Oxford University Press, 2014.

\item M.~Geva, R.~Schuster, J.~Berant and O.~Levy.
Transformer feed-forward layers are key-value memories.
In \emph{Proceedings of EMNLP}, pp.~5484--5495, 2021.

\item A.~Heuillet, F.~Cautis and N.~Talasila.
Explainability in deep reinforcement learning.
\emph{Knowledge-Based Systems}, 202:106192, 2020.

\item P.~Hohenecker and T.~Lukasiewicz.
Ontology reasoning with deep neural networks.
\emph{Journal of Artificial Intelligence Research}, 69:503--540, 2020.

\item G.~Hooker and C.~Hooker.
Machine learning and the future of realism.
arXiv preprint arXiv:1704.04688, 2017.

\item K.~Krippendorff.
\emph{Content Analysis: An Introduction to Its Methodology}.
Sage, 4th edition, 2019.

\item J.~Ladyman and D.~Ross.
\emph{Every Thing Must Go: Metaphysics Naturalized}.
Oxford University Press, 2007.

\item K.~Li, O.~Patel, F.~Viégas, H.~Pfister and M.~Wattenberg.
Inference-time intervention: Eliciting truthful answers from a language model.
In \emph{Advances in Neural Information Processing Systems}, 2023.

\item Z.~C.~Lipton.
The mythos of model interpretability.
In \emph{Proceedings of the ICML Workshop on Human Interpretability in Machine Learning}, 2016.

\item Z.~C.~Lipton.
The mythos of model interpretability.
\emph{Communications of the ACM}, 61(10):36--43, 2018.

\item P.~Mayring.
\emph{Qualitative Content Analysis: Theoretical Foundation, Basic Procedures and Software Solution}.
Beltz, 2014.

\item C.~Olah, A.~Cammarata, L.~Schubert et al.
Zoom in: An introduction to circuits.
\emph{Distill}, 5(3):e00024.001, 2020.

\item M.~J.~Page, J.~E.~McKenzie, P.~M.~Bossuyt et al.
The PRISMA 2020 statement: An updated guideline for reporting systematic reviews.
\emph{BMJ}, 372:n71, 2021.

\item B.~Russell.
\emph{The Analysis of Matter}.
Kegan Paul, 1927.

\item S.~Shapiro.
\emph{Philosophy of Mathematics: Structure and Ontology}.
Oxford University Press, 1997.

\item N.~Shea.
\emph{Representation in Cognitive Science}.
Oxford University Press, 2018.

\item B.~C.~van Fraassen.
\emph{Scientific Representation: Paradoxes of Perspective}.
Oxford University Press, 2008.

\item J.~Worrall.
Structural realism: The best of both worlds?
\emph{Dialectica}, 43(1--2):99--124, 1989.

\item M.~D.~Zeiler and R.~Fergus.
Visualizing and understanding convolutional networks.
In \emph{European Conference on Computer Vision}, pp.~818--833. Springer, 2014.

\end{enumerate}

\end{document}